\documentclass{article}
\usepackage{spconf,amsmath,graphicx,booktabs,url}
\usepackage{setspace}
\title{Speech Synthesis as augmentation for low-resource ASR}
\name{Deblin Bagchi\textsuperscript{*}\thanks{Equal contribution}, Shannon Wotherspoon\textsuperscript{*}, Zhuolin Jiang, Prasanna Muthukumar\sthanks{Corresponding author: Prasanna Muthukumar, pmuthuku@bbn.com}}
\email{\{dbagchi, swothers, zjiang, pmuthuku\}@bbn.com}
\address{Raytheon BBN, Cambridge, MA-02138, USA}

\begin{document}
%
\maketitle
\begin{abstract}
  Speech synthesis might hold the key to low-resource speech recognition. Data augmentation techniques have become an essential part of modern speech recognition training. Yet, they are simple, naive, and rarely reflect real-world conditions. Meanwhile, speech synthesis techniques have been rapidly getting closer to the goal of achieving human-like speech. In this paper, we investigate the possibility of using synthesized speech as a form of data augmentation to lower the resources necessary to build a speech recognizer. We experiment with three different kinds of synthesizers: statistical parametric, neural, and adversarial. Our findings are interesting and point to new research directions for the future. 
\end{abstract}
\section{Introduction}
\label{sec:intro}
The modern speech processing system has a voracious appetite for data. The general trend in automatic speech recognition (ASR) systems today has been to create algorithms that can make use of increasingly larger datasets. These efforts have undoubtedly yielded impressive results. The ability to learn effectively from large datasets has directly corresponded with recognition accuracy.
Far from being unique to speech recognition, ASR's counterpart, speech synthesis has also followed a similar trend. The original Blizzard challenge dataset \cite{black2005blizzard} consisted of a one-hour corpus. The most recent version \cite{blizzard2020} uses corpora several times larger.

While this trend of ever larger datasets has led to great advances in speech processing abilities, it has also ignored the needs of several end-users. Large sized corpora are simply not available for a large variety of tasks. The most common use case with this constraint is speech recognition when dealing with non-mainstream languages. Igbo, for instance, is spoken by 47 million speakers but no large corpus exists, and creating a several thousand hour Igbo corpus for ASR will be prohibitively expensive. Arguably, older techniques like Hidden Markov Models \cite{rabiner1986introduction} would still be applicable when dealing with smaller corpora. However, it is important that we do not exclude poorly resourced languages from enjoying the advances of modern speech processing technologies.

In this paper, we will therefore pursue a goal directly \emph{opposite} the general trend of papers today. We will attempt to reduce the amount of data required to build a viable speech recognizer while still using modern neural network based ASR techniques. We attempt this using speech synthesis techniques \cite{zen2009statistical} as a form of data augmentation. The reasoning behind this decision is that synthesizers typically require orders of magnitude less data than recognizers do. After all, statistical parametric speech synthesizers have even been built on as little as 30 minutes of speech data \cite{black2015random}. Speech synthesizers also offer a means of data augmentation that is less naive compared to standard augmentation techniques. For instance, a common form of data augmentation is to speed up or slow down existing speech. While this technique might improve accuracy, it does not reflect the true way that humans speed up or slow down their speech. A synthesis-based augmentation technique, on the other hand, is more likely to reflect rate changes more accurately. We therefore believe that synthesis-based augmentation should yield even better results. 

We will investigate three different forms of speech synthesis and their performance when used as a means of data augmentation for ASR. The first is the classic statistical parametric speech synthesis \cite{zen2009statistical}. The second is the popular neural synthesizer, Tacotron2 \cite{shen2018natural}. The third is the adversarial synthesizer, WGANSing \cite{chandna2019wgansing}. We report our experiences with all three synthesizers.

\section{Prior Work}
\label{sec:prior}
Low-resource ASR and TTS have a rich and varied history. Technically, any ASR paper from 10 years ago or earlier can be considered a viable approach for low-resource ASR. However, the closest related work to ours is the paper from Rosenberg et al \cite{rosenberg2019speech}. The authors describe an approach where the Tacotron speech synthesizer is used to augment real speech with the goal of increasing lexical and acoustic diversity. The major difference between Rosenberg et al. and our work is in the quantity of data used. Keeping in mind our goal of using the least amount of data possible, we use substantially smaller subsets of the Librispeech corpus \cite{panayotov2015librispeech}. Because we restricted ourselves to using as little data as possible, we also had to resort to synthesizers beyond Tacotron that Rosenberg et al. use.

Another closely related work is LRSpeech by Xu et al \cite{xu2020lrspeech}. The authors target low-resource scenarios like we do but use a high-resource language to bootstrap ASR and TTS systems for the low-resource language. We, on the other hand, assume that no corpus is available apart from that of the low-resource language. We see our approach as complementary to that of Xu et al., and hope our approach is an interesting alternative.

\section{Experiments}
\subsection{Speech recognizer}
Our speech recognition models are trained using the BBN speech recognition system, Sage \cite{hsiao2016sage}. Sage is an extension of the popular Kaldi speech recognition toolkit \cite{povey2011kaldi}. Our ASR systems are multilingual initialized \cite{ma2017multilingual} hybrid TDNN-F models \cite{povey2018tdnnf}, trained for one epoch of lattice-free MMI (LF-MMI) and two epochs of sMBR. For language modeling, we train word-level trigram models. All our experiments in this paper will focus on the acoustic model since text data is typically easier to obtain than labeled audio. We use 100 hours of Librispeech transcripts for language model data in all cases, and vary the amount of audio data as required by the experiment.

\subsection{Statistical Parametric synthesis}
For our experiments using statistical parametric speech synthesis, we use the Clustergen synthesizer \cite{black2006Clustergen}. Clustergen uses the Festival system \cite{black1998festival} to convert text into a sequence of phonemes, and uses random forests to predict mel-frequency cepstral co-efficients that correspond to the phonemes. 

While Clustergen generally produces good quality speech, we use it differently from its original intent. Clustergen is heavily optimized towards a single speaker corpus. However, most ASR corpora are multi-speaker, and only possess a few minutes of speech for each speaker. It is impossible to build a synthesis model for each speaker with such little data. We therefore attempted to build average models for groups of speakers. We identified similar speakers by clustering i-vectors \cite{dehak2010front}. We pretended that all the speech from a particular cluster was from the same speaker, and built a Clustergen model for it. These average models did not produce high quality synthesis but resulted in some interesting findings anyway. They pointed out several flaws in Mel Cepstral Distortion (MCD), the objective metric we were using to assess synthesis quality. We discuss these flaws in section \ref{sec:Discussion}. 




\begin{table}[th]
  \caption{ASR performance when using synthetic speech as additional training data}
  \label{tab:real-plus-synth}
  \centering
  \begin{tabular}{ l r }
    \toprule
    \textbf{Training data} & \textbf{WER} \\
    \midrule
    20h real & 12.7  \\
    20h real + 20h synth & 12.6 \\
    20h real + 20h synth (MCD $<$ 5) & 12.7 \\
    20h real + 20h synth (MCD $<$ 6) & 12.7 \\
    20h real + 60h synth & 13.0 \\
    20h real + 80h synth & 12.8 \\
    \midrule
    40h real & 12.0  \\
    40h real + 60h synth & 13.0 \\
    \midrule
    80h real & 11.4  \\
    80h real + 20h synth & 11.5 \\
    \bottomrule
  \end{tabular}
\end{table}

\begin{table}[th]
  \caption{ASR performance when training on either purely real or purely synthetic speech}
  \label{tab:real-vs-synth}
  \centering
  \begin{tabular}{ l r }
    \toprule
    \textbf{Training data} & \textbf{WER} \\
    \midrule
    80h real & 11.4  \\
    80h synth, unclustered, 50 voices & 36.2 \\
    80h synth, clustered, 5 voices & 47.1 \\
    \bottomrule
  \end{tabular}
\end{table}

Table~\ref{tab:real-plus-synth} shows the results of using Clustergen as a data augmentation system. Here, Clustergen was trained on the specified amount of real speech and then used to synthesize the specified amount of synthetic speech. The `MCD $<$ 5' and `MCD $<$ 6' rows correspond to experiments where we left out synthetic voices that had MCD scores higher than the number. 

Table~\ref{tab:real-vs-synth} shows the results of an experiment where we trained on either purely real speech or purely synthetic speech. The results in the second row were generated by building a separate Clustergen model for each speaker no matter how little the available data. The third row in the table is the result of clustering similar speakers using i-vectors. The results in both tables indicate that Clustergen's speech quality was not sufficient to be useful as a data augmentation method. 

For neural TTS and adversarial TTS, we used Clustergen as a baseline. Since ASR training is computationally expensive, we did not train the ASR system whenever any synthesizer performed worse than the parametric synthesizer. 


\subsection{Neural TTS}
Tacotron2 \cite{shen2018natural} is a purely neural speech synthesizer that consists of a recurrent sequence-to-sequence neural network that takes as input character embeddings and predicts mel spectrograms. We use Waveglow \cite{prenger2019waveglow} as a vocoder to convert the mel spectrograms to raw speech. Tacotron2 has been shown to produce excellent quality speech in the past, but our primary reason for choosing this synthesizer and waveglow was because there was open-source code available. 

We were successful in producing high quality synthesis from Tacotron2 by training on the 24 hour LJ Speech dataset \cite{ljspeech17}, but had difficulty achieving similar results on any 1 hour CMU Arctic dataset \cite{kominek2004cmu}. Despite our best efforts, the best Tacotron2 could do on the one-hour corpus was babble in a voice that strongly resembled the original speaker. This test is important because CMU Arctic is still larger and cleaner than any single voice we had for our ASR datasets. We therefore had to rule out any possibility of training a Tacotron2 model exclusive for each voice in an ASR corpus. (Building an exclusive model for each voice in an ASR corpus would have been terribly expensive computationally anyway). 

To overcome this difficulty, we extended Tacotron2 to support multi-speaker training by incorporating speaker embeddings as input in addition to the character embeddings. We experimented with both one-hot embedddings as well as i-vectors. One-hot embeddings gave us marginally better results than i-vectors but neither could match up to even the quality of Clustergen. Nor was there an easy way of quantifying the difference in quality between the i-vectors and one-hot embeddings. Since the quality of synthesis on our corpus was significantly worse than using classical parametric synthesis, we did not attempt our augmentation experiments.

\subsection{Adversarial TTS}
In addition to the classical parametric and neural speech synthesizers, we also tested an adversarial synthesizer based on Generative Adversarial Networks (GANs). Adversarial learning is a new machine learning technique that has had tremendous success in generating high-dimensional data. These techniques are still in the nascent stages for speech synthesis, but nevertheless show great promise because of the success they have had in related fields such as image generation. 

The toolkit we used for our experiments was actually a GAN-based singing synthesizer called WGANSing \cite{chandna2019wgansing}. We chose WGANSing because it was the only open-source toolkit we could find capable of open-vocabulary speech synthesis. The architecture of this system is the same as DCGANs \cite{blaauw2019data} and is trained using the Wasserstein GAN algorithm. 

WGANSing takes as input a block of frame-wise linguistic features and singer identity features, and outputs vocoder features that correspond to the block. The most problematic for us among the input features were the precise phone durations and the pitch contours required. These features make sense for synthesizing singing, but are very difficult to generate for regular speech. We were therefore forced to modify the toolkit to avoid conditioning on these troublesome features. However, the resulting synthesis quality without these features was poor compared to the other synthesizers discussed in this paper. We suspect that this sub-par performance is caused by the low amounts of training data we use. Low-resources are a requirement for our task. Unfortunately, this requirement might also be the Achilles heel of adversarial ML.

\section{Discussion}\label{sec:Discussion}
One major setback we faced in our series of experiments is the failure of Mel Cepstral Distortion (MCD) as a metric. In standard speech synthesis experiments, MCD has issues but is generally reliable. For instance, a rule of thumb is that an MCD reduction of 0.12 is equivalent to doubling the amount of training data \cite{kominek2008synthesizer}. Unfortunately, intuitive rules and beliefs such as these only ended up being true for the standard speech synthesis training scenario of clean, single-speaker data. When training on noisy, multi-speaker data, not only was MCD a poor reflection of quality, but it was also extremely misleading. In informal tests, we found several instances where the metric indicated high MCD but the synthesized speech was understandable. We also found several examples where the synthesized speech was unintelligible but the measured MCD was low. Clearly, single-speaker objective metrics do not extrapolate well to multi-speaker training scenarios. 

Another major pain point for us was the gordian knot that was Tacotron2. We faced tremendous engineering and research challenges in getting sensible performance from this synthesizer. The engineering challenges were primarily related to the amount of compute required (at least the implementation we used required this). Both Tacotron2 and Waveglow could only be run on the fastest GPUs we had available, and then still took between days to weeks to converge. While we managed to overcome these through a lot of work, the research challenges proved more insurmountable. We were able to successfully replicate the default experiments in the open-source Tacotron2 toolkit, but we had little success when running the codebase on our own smaller dataset. The inscrutable nature of modern neural networks also made it difficult to extend the system to support multiple speakers. Despite being experts in speech synthesis and neural networks, progress was extremely difficult. We hope that future work by us and others relieves researchers of this burden, and makes it easier to achieve positive results. 

\section{Conclusion}
The results of our experiments have mostly been disappointing. It appears that modern speech synthesis has not yet advanced to be of sufficient use in training low-resource speech recognizers. Nevertheless we still believe that as speech synthesizers increase in quality, this approach will eventually become viable. We also hope that our experiments point to interesting future directions with a higher potential for success. For instance, every synthesizer we explored in our paper has been heavily tuned for the human ear. It is highly likely that new synthesizers will need to be designed and built with the explicit goal of acting as sources of augmented data. 

\section{Acknowledgement}
We thank Sean Colbath, Ilana Heintz, and Ron Coleman for all their support in this endeavor. 


\vfill\pagebreak

\bibliographystyle{IEEEbib}
{\footnotesize
\bibliography{refs}}

\end{document}